\documentclass{article}

\usepackage{microtype}
\usepackage{graphicx}
\usepackage{booktabs} 

\usepackage{hyperref}
\usepackage{url}

\usepackage{graphicx}
\usepackage{doi}
\usepackage{caption}
\usepackage{subcaption}

\usepackage{hyperref}

\usepackage[accepted]{icml2023}

\usepackage{amsmath}
\usepackage{amssymb}
\usepackage{mathtools}
\usepackage{amsthm}

\usepackage[capitalize,noabbrev]{cleveref}

\theoremstyle{plain}

\theoremstyle{definition}

\theoremstyle{remark}

\usepackage[textsize=tiny]{todonotes}

\icmltitlerunning{GCI: A (G)raph (C)oncept (I)nterpretation Framework}

\begin{document}

\twocolumn[
\icmltitle{GCI: A Graph Concept Interpretation Framework}



\icmlsetsymbol{equal}{*}

\begin{icmlauthorlist}
\icmlauthor{Dmitry Kazhdan}{camb}
\icmlauthor{Botty Dimanov}{tenyks}
\icmlauthor{Lucie Charlotte Magister}{camb}
\icmlauthor{Pietro Barbiero}{camb}
\icmlauthor{Mateja Jamnik}{camb}
\icmlauthor{Pietro Lio}{camb}
\end{icmlauthorlist}

\icmlaffiliation{camb}{Department of Computer Science and Technology, The University of Cambridge, Cambridge, United Kingdom}
\icmlaffiliation{tenyks}{Tenyks, Cambridge, United Kingdom}

\icmlcorrespondingauthor{Dmitry Kazhdan}{dk525@cam.ac.uk}

\icmlkeywords{Machine Learning, ICML}

\vskip 0.3in
]



\printAffiliationsAndNotice{\icmlEqualContribution} 

\begin{abstract}
Explainable AI (XAI) underwent a recent surge in research on concept extraction, focusing on extracting human-interpretable concepts from Deep Neural Networks. An important challenge facing concept extraction approaches is the difficulty of interpreting and evaluating discovered concepts, especially for complex tasks such as molecular property prediction. We address this challenge by presenting GCI: a (G)raph (C)oncept (I)nterpretation framework, used for quantitatively measuring alignment between concepts discovered from Graph Neural Networks (GNNs) and their corresponding human interpretations. GCI encodes concept interpretations as functions, which can be used to quantitatively measure the alignment between a given interpretation and concept definition. We demonstrate four applications of GCI: (i) quantitatively evaluating concept extractors, (ii) measuring alignment between concept extractors and human interpretations, (iii) measuring the completeness of interpretations with respect to an end task and (iv) a practical application of GCI to molecular property prediction, in which we demonstrate how to use chemical \textit{functional groups} to explain GNNs trained on molecular property prediction tasks, and implement interpretations with a $0.76$ AUCROC completeness score.
\end{abstract}

\section{Introduction}

Graph Neural Networks (GNNs) have emerged as a powerful paradigm for Deep Learning on graphs, and have been applied in a wide variety of domains including social, biological, physical, and chemical \citep{sanchez-lengeling2021a, zhou2020graph, wu2020comprehensive, scarselli2008graph}. Similarly to other Deep Learning approaches, such as those based on Convolutional Neural Networks (CNNs), GNNs are black-box models whose behaviour cannot be interpreted directly. This lack of explainability hinders deployment of such types of models, especially in safety-critical applications \cite{ying2019gnnexplainer}.

Consequently, recent research has attempted to address this issue by applying existing Explainable AI approaches to GNNs \cite{yuan2022explainability, pope2019explainability, ying2019gnnexplainer}. The most widely used explainability approaches for applied to GNNs are feature importance methods \cite{ying2019gnnexplainer}. For a given data point, these methods provide scores that show the importance of each feature (i.e., subgraph, node, or edge) to the algorithm's decision. Unfortunately, these methods have been shown to be fragile to input~\cite{ghorbani2017interpretation,kindermans2019reliability} or model parameter~\cite{adebayo2018sanity,dimanov2020you} perturbations. Human experiments also demonstrate that feature importance explanations do not necessarily increase human understanding, trust, or ability to correct mistakes in a model~\cite{poursabzi2018manipulating,tcav}.

As a consequence, more recent approaches to GNN explainability have focused on \textit{concept-based} explanations~\citep{magister2021gcexplainer, magister2022encoding, georgiev2022algorithmic}. Concept-based approaches aim to provide explanations of a Deep Neural Network (DNN) model in terms of human-understandable units, rather than individual features, pixels, or characters~\cite{zhou2018interpretable,ghorbani2019towards, dimanov2021interpretable, zarlengaconcept, kazhdan2020meme}. In particular, \textit{concept extraction} approaches (a subfield of concept-based explanations) aim to extract interpretable concepts from Deep Learning models, in order to better understand which concepts a model has learned, and see if a model has picked up novel concepts that could improve existing domain knowledge \cite{ghorbani2019towards, magister2021gcexplainer}.

A key challenge for concept extraction approaches is that they are predominantly evaluated using qualitative methods, such as visually inspecting whether an extracted concept contains groups of superpixels that are semantically meaningful (e.g., `these superpixels all contain a wheel, therefore the likely concept is \textit{wheel}')~\citep{kazhdan2020cme,ghorbani2019towards,kim2017interpretability, zarlengaconcept}. However, this qualitative approach makes it difficult to assess which interpretations are more/less accurate. Furthermore, this makes it difficult to compare different concept extraction approaches to each other. 

We address the above challenge by presenting GCI: a (G)raph (C)oncept (I)nterpretation framework\footnote{Code is available at https://github.com/dmitrykazhdan/gci}, used for \textit{quantitatively} verifying concepts extracted from GNNs, using provided human interpretations. Specifically, GCI encodes user-provided concept interpretations as functions, which can be used to quantitatively measure alignment between extracted concepts and user interpretations. 


In summary, we make the following contributions:

\begin{enumerate}
    \item{We present GCI: (G)raph (C)oncept (I)nterpretation, the first framework capable of quantitatively analysing extracted concepts for GNNs on graph classification tasks;}
    
    \item{Using several synthetic case-studies, we demonstrate how GCI can be used for (i) quantitatively evaluating the quality of concept extractors, (ii) measuring alignment between different concept extractors and human interpretations, (iii) measuring the completeness of interpretations with respect to an end task;}
    
    \item{We demonstrate how GCI can be used for molecular property prediction tasks, by using a Blood-Brain Barrier Penetration task, and encoding \textit{functional groups} as concepts. In particular, we show how GCI can be used to explain a GNN model trained on this molecular property prediction task, and implement interpretations achieving a $0.76$ AUCROC completeness score.
    }

\end{enumerate}


\section{Related Work} \label{sec:related-work}

\textbf{Concept Extraction} Existing concept extraction approaches have primarily been studied in the field of Computer Vision. These approaches extract concepts from a CNNs hidden space in an \textit{unsupervised} fashion, representing these concepts as groups of superpixels~\citep{kazhdan2020cme,dimanov2021interpretable,ghorbani2019towards,kim2017interpretability, yeh2020completeness, zarlengaconcept}. More recently, variants of these approaches have been applied to GNNs as well, where concepts are extracted in the form of subgraphs in an unsupervised fashion~\citep{magister2021gcexplainer, magister2022encoding}. In this work, we rely on GCExplainer \citep{magister2021gcexplainer} as the primary benchmark in our experiments.

\textbf{Concept-based Explanations \& Graph Neural Networks} Concept-based explanations have been applied to GNNs in two primary forms: for performing concept extraction from GNN models \citep{magister2021gcexplainer, xuanyuan2022global}, and for using concepts to build GNN models \textit{interpretable by design}~\citep{magister2022encoding, georgiev2022algorithmic}. Importantly, similarly to computer vision, in all these cases the concepts are either assumed to be known beforehand, or are interpreted via visual inspection. Our GCI framework builds on top of these approaches by allowing quantitative interpretation of potentially-novel concepts.

\textbf{Concept Verification} 
Several methods have been proposed for evaluating concept extraction approaches, besides using visual inspection. Work by \cite{ghorbani2019towards} and \cite{yeh2020completeness} relies on human studies to evaluate properties such as \textit{concept coherence}. Work by \cite{magister2021gcexplainer} introduces \textit{concept purity} for measuring the difference between graphs of a given concept, which relies on computing graph edit distance. Work in~\cite{zarlengaquality} propose mutual information based metrics to evaluate quality of concept representations. Finally, work by \cite{finzel2022generating} introduces a validation framework for GNN explainers, in which a set of Inductive Logic Programming rules are produced from domain experts and GNN explanations. These are then compared with each other for validation. Importantly, these approaches either rely on time-intensive user studies, or metrics which are computationally-infeasible for a large number of graphs (such as edit distance), or assume the concepts have already been interpreted and labelled (as in~\cite{zarlengaquality}), or are applicable to niche types of graph datasets (as in the last case). In contrast, GCI is fast to setup, computationally feasible, is used for \textit{intepreting} concepts, and is applicable to any type of graph data.


\section{Methodology} \label{sec:methodology}

\begin{figure*}[h!]
    \centering
    \includegraphics[width=0.95\textwidth]{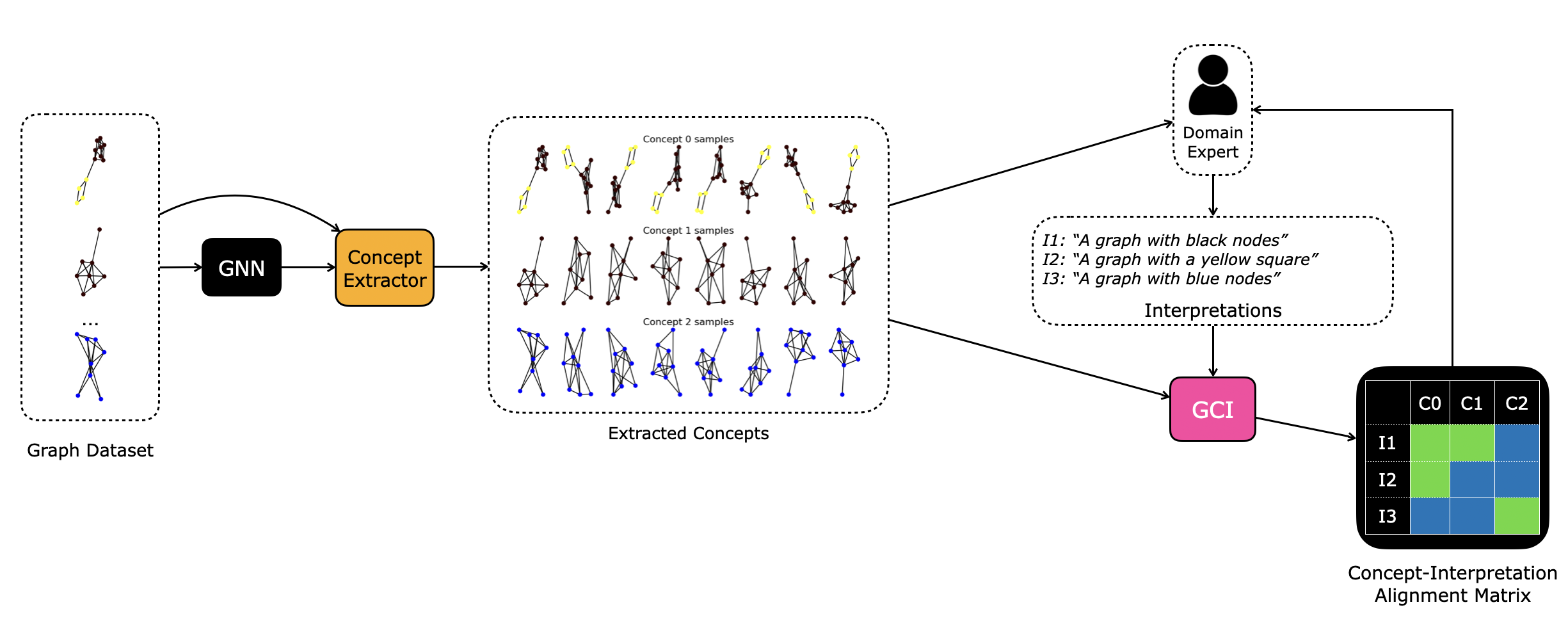}
    \caption{GCI framework overview. Left-to-right: Firstly, a Concept Extractor extracts a set of concepts from a GNN trained on a graph dataset. Next, these concepts are observed by a domain expert. The domain expert then generates a set of \textit{interpretations} of the generated concepts. These interpretations and concepts are fed into the GCI framework, which generates an Interpretation Alignment matrix, showing the degree to which the interpretations align with the concepts. The domain expert can then use this information to further refine and/or improve the produced interpretations, and consequently the domain knowledge, by better understanding the extracted concepts.} 
    \label{fig:vis_abs}
\end{figure*}

In this section we present our GCI approach, describing how it can be used for quantitatively analysing extracted concepts. A visual summary of our GCI approach is shown in Figure \ref{fig:vis_abs}. 

\subsection{Concept Extraction}

Existing work on Concept Extractors (CEs) typically represents a concept as a set of images/super-pixels (in case of computer vision), or as set of graphs in case of GNNs. More formally, we can therefore represent a concept extraction approach $f_{ext}$ as a function $f : (\mathcal{G}, \mathcal{M}) \to \mathcal{P}$, taking in a set of graphs $G \in \mathcal{G}$, together with a GNN model $M \in \mathcal{M}$ trained on this graph set, and returning a set of sets of graphs (i.e., the subset of the powerset of $\mathcal{G}$, defined as $\mathcal{P}$). Hence, $f_{ext}(G, M) = \{ G_{1}, ..., G_{m} \}$ maps a given graph set $G$ (the model's \textit{training dataset}) and model $M$ to a set of sets of graphs $\{ G_{1}, ..., G_{m} \}$, where each set $G_{i}$ represents a concept $C_{i}$.  


\subsection{Concept Interpretations}

Once a CE has returned a set of concepts, the next step involves \textit{interpreting} the extracted concepts, and attempting to describe what they may represent (e.g., `I think this concept represents graphs with blue nodes'). More formally, this process may be seen as defining a set of \textit{interpretation functions} $h : \mathcal{Z} \to \{0, 1 \}$. Here, $h(g)$ returns \textit{true} or \textit{false}, depending on whether the given anticipated property is or is not present in a given graph $g$ (e.g., `this is a graph with blue nodes').

\subsection{Interpretation-Concept Alignment}
Intuitively, a `good' interpretation for a concept is one which generally holds for all or most examples of that concept. For example, an interpretation of `this concept represents graphs with blue nodes' can be considered good, if graphs of this concept are indeed predominantly graphs with blue nodes. Using the above definitions, we can measure the \textit{alignment} between a given concept and an interpretation as $q : \mathcal{G} \times \mathcal{H} \to \mathbb{R}$, where $q(G, h)$ is defined as:

\begin{equation}
    q(G, h) = \frac{|\{ h(g) \ \text{is true}, \ g \in G \}|}{|G|} 
\end{equation}

Hence, the alignment between concept samples $G$ and an interpretation $h$ is the fraction of the samples for which $h$ is true, that is, the \textit{precision} of $h$ with respect to $G$. Consequently, given a set of concept samples $C=\{ G_{1}, ..., G_{m} \}$, and a set of interpretations $H=\{ h_{1}, ..., h_{n} \}$, we define an \textit{Interpretation-Alignment} (IA) matrix as:

\begin{equation}
    IA(C, H)_{i, j} = q(G_{i}, h_{j}), \forall i, j.
\end{equation}

\subsection{Further Interpretation Metrics}

The above definition for interpretations lends itself to numerous further metrics that can be built on top of interpretations. For instance, we can introduce the notion of `interpretation hierarchies', by seeing if one interpretation is a `subset' of another, that is, $h_{1} \subset h_{2} \Leftrightarrow \forall g \in \mathcal{Z}: h_{1}(g) \Rightarrow h_{2}(g)$. Alternatively, we can also use this definiton to quantitatively explore relationships between interpretations by measuring their mutual information. We focus on the precision-based IA matrix in this work, leaving extensions of these metrics for future work.

\textbf{Interpretation Completeness \& Predictability} In this work, we also build on top of the \textit{concept completeness} metric introduced in \citep{yeh2020completeness}, and \textit{concept predictability} metric introduced in \cite{kazhdan2020cme}. The completeness of a set of concepts is defined as the accuracy of a classifier with respect to the end task, which uses \textit{only} concept information as input (i.e., `how predictable are the end-task labels from the concept information alone?') \citep{yeh2020completeness}. The predictability of a given concept from a given model is defined as the accuracy of a classifier trained to predict the values of that concept from the model's hidden space (further details are given in \cite{kazhdan2020cme}).

We build on top of the above notions, by observing that we can apply our interpretation functions \textit{directly} to a model's training graph dataset, and thus obtain an \textit{interpretation representation} of this dataset. We can then measure the completeness and predictability of this representation, in the same way as the completeness and predictability of a concept representation. We refer to these metrics as \textit{interpretation completeness} and \textit{interpretation predictability} in the remainder of this work. Intuitively, these metrics represent how \textit{predictable} human-provided interpretations are from a given GNN (which serves as a proxy for how much the model learned this information), as well as how \textit{complete} these interpretations are with respect to an end task (i.e. how well do these interpretations describe the actual task). Collectively, these metrics demonstrate how \textit{relevant} interpretations are for the end task \cite{kazhdan2020cme}.


\section{Experiments \& Results}

In this section, we demonstrate the utility of the GCI framework on a range of synthetic baselines, as well as on a practical use-case of molecular property prediction. Similarly to \cite{ying2019gnnexplainer}, we rely on synthetically-generated Barabási-Albert graphs for our synthetic baseline experiments.

\subsection{Concept Extractor Evaluation}

Firstly, we demonstrate how GCI can be used to quantitatively evaluate different CEs of varying quality, and compare it with a visual inspection approach. In this setup, we relied on a synthetic graph classification dataset with a trivial concept structure, and simulated a reduction in CE quality by injecting random perturbations into its output.

\textbf{Dataset:} We generated $1000$ Barabási-Albert graphs. $50$\% of these graphs were coloured blue, $25$\% were coloured red, and $25$\% were coloured yellow. The blue graphs were assigned class label $0$, whilst the red and yellow graphs were assigned class label $1$. Overall, this dataset represents a graph classification task consisting of $2$ classes, in which class $1$ is composed of 2 distinct concepts (yellow and red node colours).

\textbf{GNN Model:} We trained a standard Graph Convolutional Network \citep{zhang2019graph} on the above binary graph classification task, achieving $100\%$ accuracy.

\textbf{Concept Extractor Baselines:} We used 3 different baselines for our CEs: \textit{GCExplainer}, \textit{NoisyGCExplainer}, and \textit{RandomExplainer}. GCExplainer uses the methodology described in \cite{magister2021gcexplainer}. NoisyGCExplainer works by taking the output of a GCExplainer, and randomly moving a fraction $\theta$ of the subgraphs from one concept to another (for this experiment, we set $\theta=20\%$). This baseline represents a more noisy, imperfect CE. Finally, RandomExplainer takes the output of a GCExplainer, and randomly shuffles the subgraphs across the concepts, representing a baseline random concept predictor. We assumed the \textit{number} of concepts is known in these experiments, and set that to $3$ for the GCExplainers. Overall, these baselines represent CEs of varying quality, in which a reduction in quality is achieved by injecting randomness into concept outputs. Further details are given in Appendix \ref{appendix:random_gcexplainer}.

\textbf{Interpretations:} We provided 3 \textit{ground truth} interpretations, defined as $h_{0}(g) \Leftrightarrow \text{`g is a graph with red nodes'}$, $h_{1}(g) \Leftrightarrow \text{`g is a graph with blue nodes'}$, $h_{2}(g) \Leftrightarrow \text{`g is a graph with yellow nodes'}$.

\begin{figure*}[t!]
    \centering
    \begin{subfigure}[b]{0.3\textwidth}
    \centering
        \includegraphics[width=\textwidth]{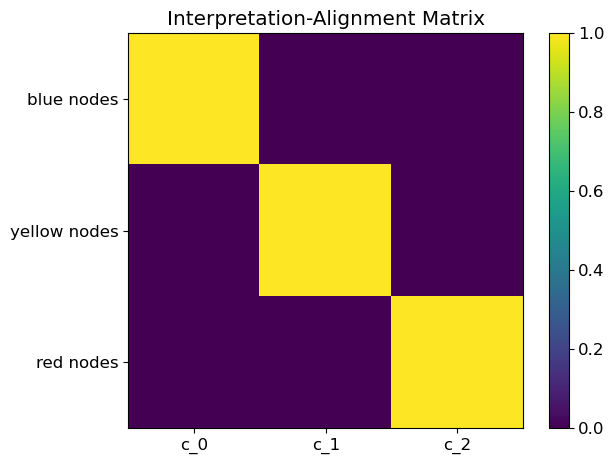}
        \caption{GCExplainer}
        \label{fig:same_accuracy_GCExplainer}
    \end{subfigure}
    \hfill
    \begin{subfigure}[b]{0.3\textwidth}
        \centering
        \includegraphics[width=\textwidth]{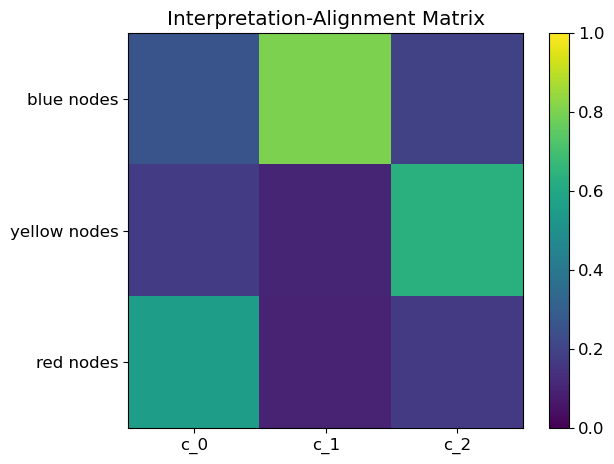}
        \caption{Noisy GCExplainer}
        \label{fig:SameAccuracy}
    \end{subfigure} 
    \hfill
    \begin{subfigure}[b]{0.3\textwidth}
        \centering
        \includegraphics[width=\textwidth]{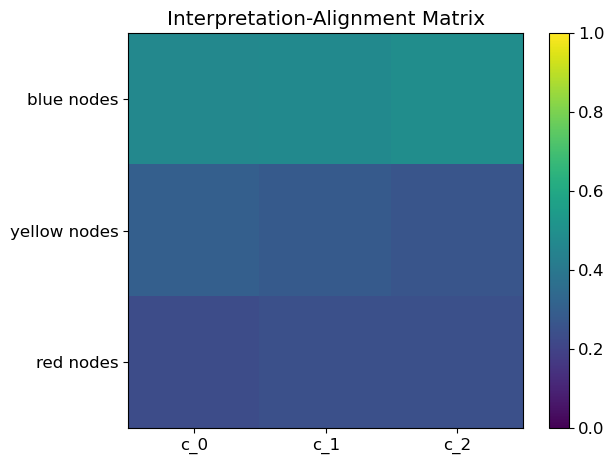}
        \caption{Random Explainer}
         \label{fig:SameAccuracy}
    \end{subfigure} 
    \caption{IA-Matrices for the different heuristics. Each column corresponds to an extracted concept ($c_0$-$c_2$), and each row corresponds to a defined interpretation ($h_0$-$h_2$). Each matrix cell $(i, j)$ represent the alignment between interpretation $h_i$ and concept $c_j$. As can be seen - each GCExplainer concept achieves high alignment with one of the ground truth interpretations, indicating that GCExplainer has successfully recovered the underlying concepts. Other concept extractors achieve significantly lower alignment.}
    \label{fig:synth_extractor_eval_matrix}
\end{figure*}

\begin{figure*}[t!]
    \centering
    \begin{subfigure}[b]{0.3\textwidth} 
    \centering
        \includegraphics[width=\textwidth]{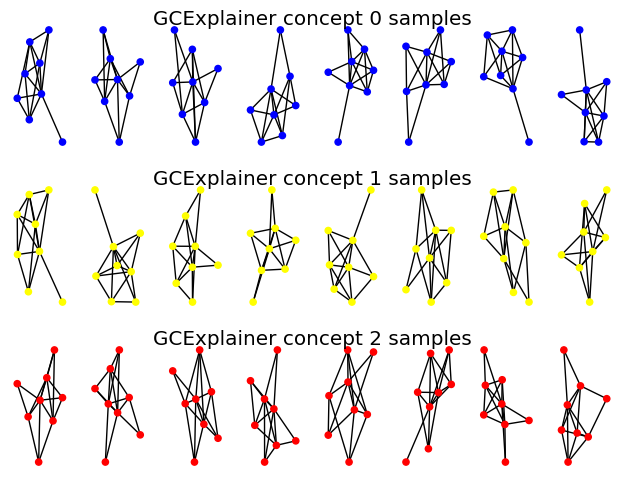}
        \caption{GCExplainer}
    \end{subfigure}
    \hfill
    \begin{subfigure}[b]{0.3\textwidth}  
        \centering
        \includegraphics[width=\textwidth]{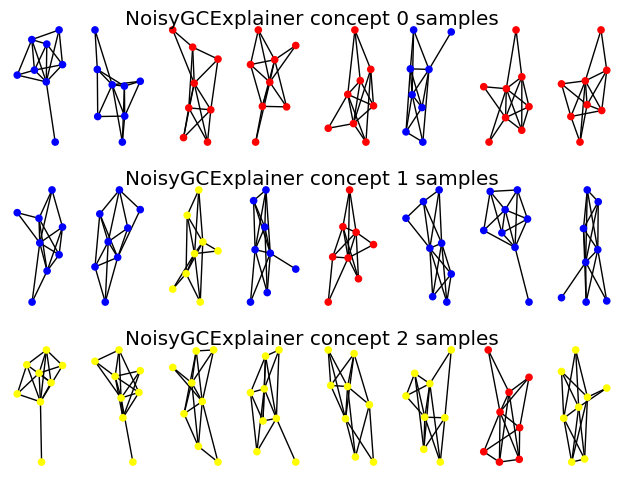}
        \caption{Noisy GCExplainer}
    \end{subfigure} 
    \hfill
    \begin{subfigure}[b]{0.3\textwidth}
        \centering
        \includegraphics[width=\textwidth]{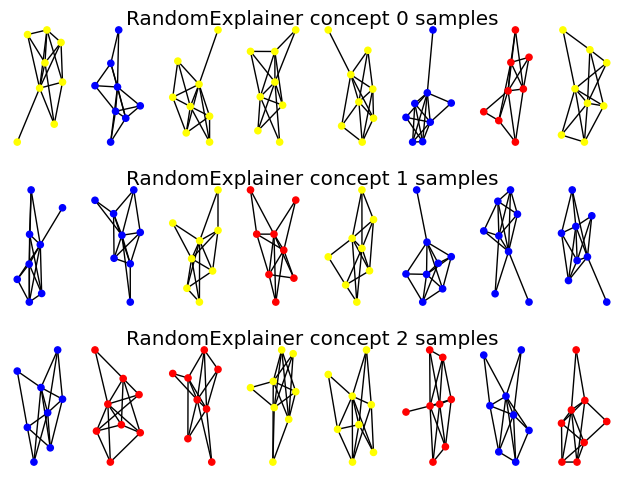}
        \caption{Random Explainer}
    \end{subfigure}
    \caption{Representative concept samples for the different explainers. Notice that each row depicts the examples contained within a single concept. The more similar the graphs within the concept, the higher the purity of the concept; hence, the explainer is better. Observe how this information is accurately summarised in the IA matrix in Figure~\ref{fig:synth_extractor_eval_matrix}.}
    \label{fig:synth_extractor_eval_graph}
\end{figure*}

\textbf{Results} The IA matrices between the different explainer concepts and interpretations are presented in Figure \ref{fig:synth_extractor_eval_matrix}. It is clear that GCExplainer has near-perfect alignment with the interpretations, whilst the other two explainers demonstrate a significantly lower alignment. Extracted concept samples are shown in Figure \ref{fig:synth_extractor_eval_graph}. Crucially, the IA matrix quantitatively shows the vast difference between explainers (with GCExplainer aligning the best). Such a conclusion would have been much more difficult to draw from simple visual inspection.

\subsection{Intepretation Alignment Grading}

In this section, we demonstrate how GCI can be used to quantitiatively measure the degree of alignment between human interpretations and extracted concepts, using a synthetic graph classification task. In particular, we rely on CEs of varying \textit{granularity} (with one CE extracting relatively more lower-level concepts), and show how GCI can be used to determine which sets of interpretations best describe which extractor. As before, we compare this to a visual inspection approach.

\textbf{Dataset:} We generated $1000$ Barabási-Albert graphs synthetically. $50$\% of these graphs were coloured blue. The remaining $50$\% of the graphs were coloured red. Furthermore, we attached a square subgraph structure to $80$\% \textit{of the red graphs}. The blue graphs were assigned class label $0$, and the red graphs a class label $1$. Overall, this dataset represents a graph classification task consisting of 2 classes, in which one of the classes has 2 highly correlated concepts (a square subgraph, and the colour red).

\textbf{GNN Model:} We trained a standard GCN on the above binary graph classification task, achieving $100\%$ accuracy.

\textbf{Concept Extractor Baselines:} We used $2$ different GCExplainer baselines: one with the number of concepts set to $2$, and another one set to $3$. These two baselines represent CEs of varying \textit{granularity}.

\textbf{Interpretations:} We provided 4 different interpretations defined as follows: \mbox{$h_{0}(g) \Leftrightarrow \text{`g is a blue graph'}$}; \mbox{$h_{1}(g) \Leftrightarrow \text{`g is a red graph'}$}; $h_{2}(g) \Leftrightarrow \text{`g is a graph with an}$ $\text{attached square subgraph'}$; \mbox{$h_{3}(g) \Leftrightarrow h_{1}(g) \text{ AND } h_{2}(g)$}. 

\begin{figure*}[t]
    \centering
    \begin{subfigure}[b]{0.3\textwidth} 
    \centering
        \includegraphics[width=\textwidth]{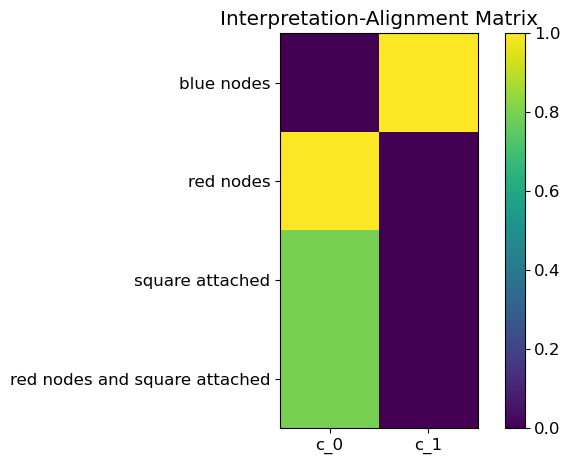}
        \caption{2-concept GCExplainer IA matrix}
        \label{fig:alignment_subfig1}
    \end{subfigure} \hspace{0.2\textwidth}
    \begin{subfigure}[b]{0.3\textwidth}  
        \centering
        \includegraphics[width=\textwidth]{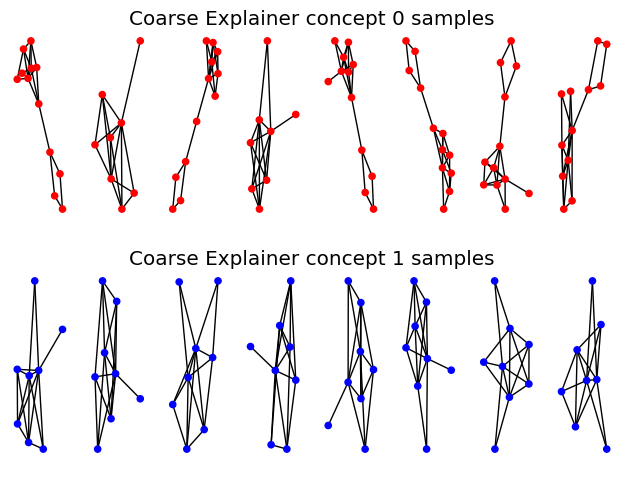}
        \caption{2-concept GCExplainer samples}
    \end{subfigure} 

    \begin{subfigure}[b]{0.3\textwidth} 
    \centering
        \includegraphics[width=\textwidth]{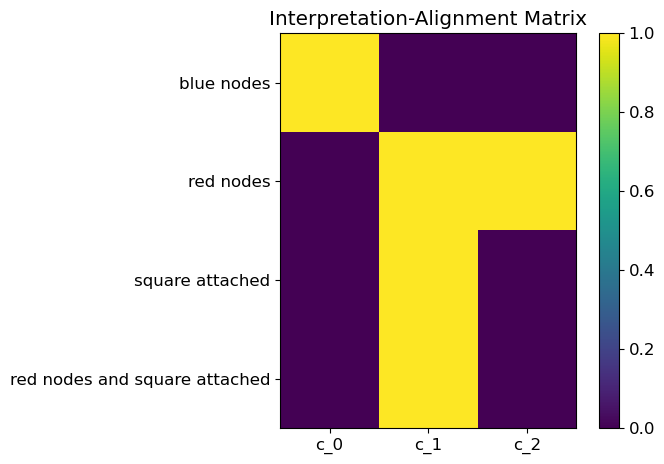}
        \caption{3-concept GCExplainer IA matrix}
    \end{subfigure} \hspace{0.2\textwidth}
    \begin{subfigure}[b]{0.3\textwidth}  
        \centering
        \includegraphics[width=\textwidth]{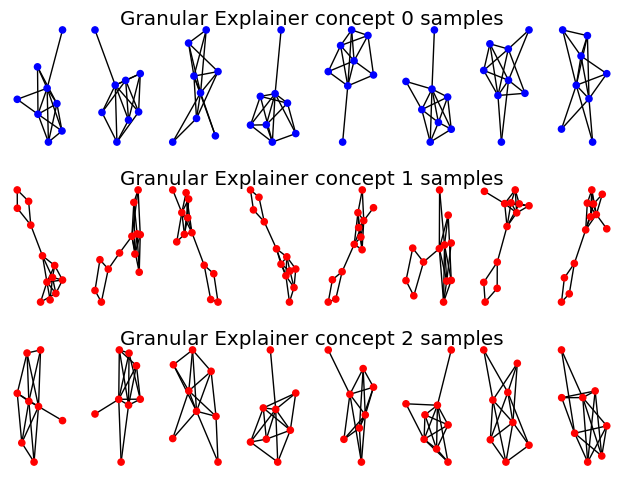}
        \caption{3-concept GCExplainer samples}
    \end{subfigure} 
    \caption{Representative concept samples (right) and corresponding IA matrices (left) for the concepts of two different explainers of varying granularity based on the number of concepts they can represent (a) \& (b) vs (c) \& (d), for 2 and 3 concepts, respectively.}
    \label{fig:synth_alignment_samples_graph}
    
\end{figure*}

\textbf{Results:} Figure \ref{fig:synth_alignment_samples_graph} shows the IA matrices and extracted concept samples for both explainers. Importantly, we observe that at lower granularity, only $h_{0}$ and $h_{1}$ accurately describe the extracted concepts, whilst at higher granularity, the CE was successfully able to separate out the two concepts of class $1$, such that all 4 heuristics accurately describe the different concepts. Such a conclusion regarding which sets of interpretations better describe the extracted concepts would have been very difficult to draw simply by observing the concept samples (shown in Figure~\ref{fig:synth_alignment_samples_graph} (b) \& (d)).

\subsection{Model-Interpretation Completeness \& Predictability}


In this section, we demonstrate how GCI can be used to measure the completeness of a set of interpretations with respect to an end task, and the predictability of these interpretations from a given model (as discussed in Section \ref{sec:methodology}). Furthermore, we demonstrate that GCI can be used to verify concepts, even if they are \textit{not} relevant to an end-task.


\textbf{Dataset:} We generated $1000$ Barabasi-Albert graphs synthetically. $50$\% of these graphs were coloured blue and assigned a class label $0$, whilst the other $50$\% were coloured red and assigned a class label $1$. For each of the classes, we randomly selected $15$\% of the samples, and coloured one randomly selected node in \textit{each} graph purple. We then randomly selected $15$\% of the samples from \textit{each} class again, and added a square subgraph to them. Overall, this dataset represents a graph classification task where there are 2 concepts directly related to the end task (the red and blue colour), and 2 concepts that are \textit{independent} of the end task (the purple node and the square structure).

\textbf{GNN Model:} We trained a standard GCN on the above binary graph classification task, achieving $99\%$ accuracy.

\textbf{Concept Extractor Baseline:} We use a GCExplainer with the number of concepts set to 4 in this example (as before, we assumed the number of concepts is known in these experiments).

\textbf{Interpretations:} We provided 4 different \textit{ground truth} interpretations, corresponding to the $4$ underlying concepts: $h_{0}(g) \Leftrightarrow \text{`g is a blue graph'}$; $h_{1}(g) \Leftrightarrow \text{`g is a red graph'}$; $h_{2}(g) \Leftrightarrow \text{`g is a graph with an attached square subgraph'}$; $h_{3}(g) \Leftrightarrow \text{`g contains a purple node'}$. 

\begin{figure*}[t]
    \centering
    \begin{subfigure}[b]{0.3\textwidth} 
    \centering
        \includegraphics[width=\textwidth]{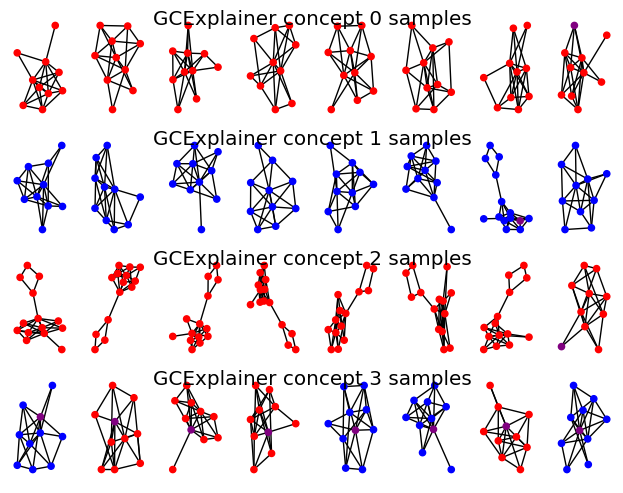}
        \caption{GCExplainer concept samples}
        \label{fig:completeness_samples}
    \end{subfigure}
    \hfill
    \begin{subfigure}[b]{0.3\textwidth}  
        \centering
        \includegraphics[width=\textwidth]{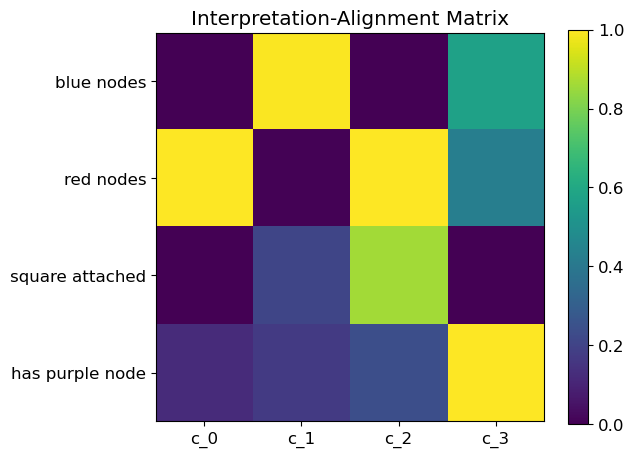}
        \caption{GCExplainer IA Matrix}
        \label{fig:completeness_matrix}
    \end{subfigure} 
    \hfill
        \begin{subfigure}[b]{0.3\textwidth}  
        \centering
        \includegraphics[width=\textwidth]{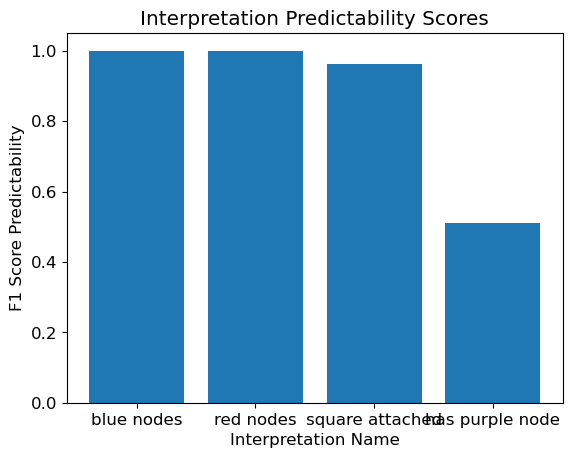}
        \caption{GCExplainer Concept Predictability}
        \label{fig:completeness_bar}
    \end{subfigure} 
    \caption{Representative concept samples (a), corresponding IA matrices (b) for the concepts, and  (c) GCExplainer Concept Predictability Scores. Notice that concepts relevant to the task have higher F1 Score predictability than irrelevant concepts. }
    \label{fig:completeness_predictability}
\end{figure*}

\textbf{Results:} Firstly, we show the concept samples from the CE in Figure \ref{fig:completeness_samples}, as well as the IA matrix in Figure \ref{fig:completeness_matrix}. Importantly, \textit{all} underlying concepts have been picked up by the CE successfully (with above $70\%$ precision).

Next, we applied our interpretation functions to the input graph dataset to obtain its interpretation representation, and computed the completeness of that representation (as discussed in Section \ref{sec:methodology}). We obtained a completeness score of $99\%$ for this interpretation representation. Furthermore, we measured the predictability of this representation, with the results shown in Figure \ref{fig:completeness_bar} (further details are given in Appendix \ref{appendix:concept_predictability}). Overall $h_{2}$ and $h_{3}$ obtained relatively lower predictability scores.

The above results demonstrate that we can use GCI to measure predictability and completeness of interpretations in the same way completeness and predictability is measured for concepts. Furthermore, the above results are consistent with the findings made in \cite{kazhdan2020now}, which showed that concepts with a lower dependence on the end-task will usually have lower predictability, since the model does not need to retain information about these concepts in order to achieve high performance.

Importantly, Figure~\ref{fig:completeness_matrix} also demonstrates that GCI and \mbox{GCExplainer} successfully extracted and verified \textit{all} underlying concepts, not just the ones directly related to the end task. This is significant, since this indicates that GCI and corresponding concept extractors are not restricted to only extracting and verifying concepts directly related to the end task.

Overall, the above results show how GCI can be used to \mbox{(i) measure} how complete a given set of interpretations is with respect to an end task, (ii) measure the degree to which a given model learns these interpretations (via predictability), (iii) verify extracted concepts using these interpretations, including concepts \textit{not} related to the end-task

\subsection{Application: Molecular Property Prediction}

In this section, we demonstrate how GCI can be applied to a practical use-case of \textit{molecular property prediction} \cite{walters2020applications}. This is achieved by integrating GCI with \textit{TorchDrug}\footnote{https://torchdrug.ai/} (a framework for drug discovery with machine learning), and using one of the available molecular prediction benchmarks from there \cite{zhu2022torchdrug, wu2018moleculenet}.

\subsubsection{Molecular Property Prediction Interpretations}

When dealing with chemistry tasks such as molecular property prediction, there are multiple ways of representing information about the underlying molecular graphs, at varying levels of granularity. For instance, a molecule can be described by its molecular mass (high-level information), by its SMILES representation \cite{weininger1990smiles}, or by its molecule type, etc.

Intuitively, selected interpretations used to describe extracted concepts should `match' the granularity of the extracted concepts in order to describe them well. It was previously shown that CEs, such as GCExplainer \cite{magister2021gcexplainer}, typically group graphs sharing similar subgraph structures. Hence, we need to rely on interpretations, that also operate on the subgraph level when describing graphs.

Consequently, we rely on chemical \textit{functional groups} as a potential source of interpretations in this work. Informally, a functional group is defined as a group of atoms responsible for the characteristic reactions of a particular compound \cite{carey2007advanced}. Hence, functional groups represent graph information at a `subgraph level', and are a suitable choice for representing interpretations. We leave the exploration of other types of interpretations for future work.

\subsubsection{Setup}

\textbf{Dataset:} We used the Blood-Brain Barrier Penetration (BBBP) dataset as our benchmark \cite{martins2012bayesian}. This benchmark consists of a binary graph classification task, where the goal is to predict whether a given molecule will penetrate the blood-brain barrier. The blood-brain barrier blocks most drugs, hormones and neurotransmitters. Hence, the penetration of the barrier forms a long-standing issue in development of drugs targeting the central nervous system \cite{wu2018moleculenet, martins2012bayesian}. 

\textbf{GNN Model:} We trained a standard GCN, achieving $0.88$ AUROC performance (consistent with state-of-the-art findings). Further details can be found in Appendix \ref{appendix:gnn_archs}.

\textbf{Concept Extractor Baseline:} We used the GCExplainer CE with different configurations for the number of concepts. We show the results of the best-performing configuration here, and include others in Appendix \ref{appendix:bbbp_concept_var}.

\textbf{Interpretations:} In this work, we implemented $6$ well-known functional groups, including: \textit{hydroxyl group}, \textit{ketone group}, \textit{phenyl group}, \textit{chlorine group}, \textit{fluorine group}, \textit{carboxyl group}. Finally, we also implemented an interpretation which checks whether a molecule has at least one \textit{aromatic ring} \cite{sandler2013organic}. This set of interpretations represents functional groups that are relatively conceptually simple. An overview of these interpretations is given in Appendix \ref{appendix:interpretations}. We leave implementation of more functional groups and other forms of chemistry heuristics for future work.

\subsubsection{Results}

\begin{figure*}[t]
    \centering
    \begin{subfigure}[b]{0.48\textwidth} 
    \centering
        \includegraphics[width=\textwidth]{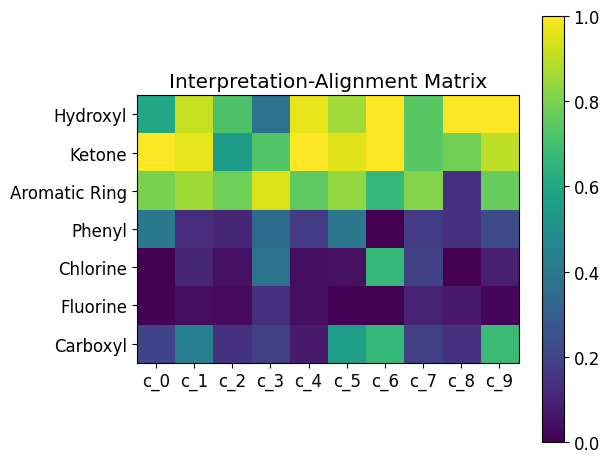}
        \caption{Class $0$ (no barrier penetration) IA Matrix.}
    \end{subfigure}
    \hfill
    \begin{subfigure}[b]{0.48\textwidth}  
        \centering
        \includegraphics[width=\textwidth]{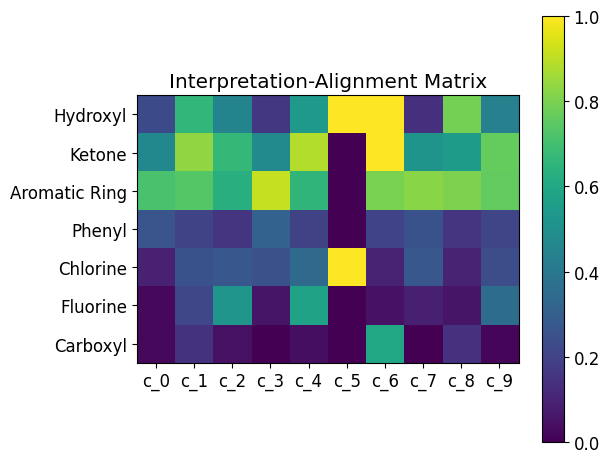}
        \caption{Class $1$ (barrier penetration) IA Matrix.}
    \end{subfigure} 
    \caption{IA matrices for the two Blood-Brain Barrier Penetration (BBBP) classes, using GCExplainer as the concept extractor, and functional groups as interpretations.}
    \label{fig:bbbp_matrix}
\end{figure*}

Firstly, we analyse concepts extracted by GCExplainer from samples of each of the classes, and show the corresponding class IA matrices in Figure~\ref{fig:bbbp_matrix}. 

Importantly, Figure \ref{fig:bbbp_matrix} demonstrates how GCI can be used to \mbox{(i) show} which interpretations are important/relevant for an end-task, and \mbox{(ii) how} these interpretations differ accross classes. In particular, we see that the top 3 interpretations (hydroxyl, ketone, and the aromatic ring) generated a large alignment with some of the extracted concepts for both classes. 

Furthermore, we can use the IA matrices to see which interpretations contribute to \textit{individual classes}. For instance, the Chlorine functional group does not resonate with any concept from class $0$, but strongly resonates with the 6th concept ($c_5$) in class $1$, together with hydroxyl. In practice, such an analysis can be a useful way to quickly visualise and interpret the key concepts contributing to individual classes.



\textbf{Completeness} Similarly to the previous experiment, we also measured the completeness of the implemented interpretations, achieving a $0.76$ AUCROC score, only $0.12$ points below the original model score of $0.88$. 

Collectively, these results demonstrate how GCI can be used to (i) encode domain knowledge heuristics (in this case - functional groups) as interpretations, (ii) use these interpretations to see which extracted concepts are aligned with them the most, (iii) verify how \textit{complete} these interpretations are, with respect to the end task. 

In the above experiment, the implemented functional groups already achieved a completeness score close to that of the original model, indicating that these interpretations serve as a good representation of the end task. We leave implementation of more types of interpretations representing the chemistry domain for future work. Further results can be found in Appendix \ref{appendix:further_results}.

\section{Conclusions}

We introduced GCI: a Graph Concept Interpretation framework, used to quantitatively interpret concepts extracted from GNNs. Using several case-studies, we demonstrate how GCI can be used to (i) quantitatively evaluate quality of different concept extractors, (ii) quantitatively measure the degree of alignment between a set of concepts \& human interpretations, (iii) measure the task completeness of human interpretations. Furthermore, we show how GCI can be applied to a realistic molecular property prediction case-study, showing how it can be used to interpret the extracted concepts. Given the rapidly-increasing interest in concept-based explanations of GNN models, we believe GCI can play an important role in mining new concepts from GNNs, as well as verifying whether or not GNNs have picked up the intended concepts.

\subsubsection*{Acknowledgments}

DK acknowledges support from the EPSRC ICASE scholarship and GSK.

\bibliography{example_paper}
\bibliographystyle{icml2023}


\appendix
\onecolumn

\section{Experimental Setup}

\subsection{Randomised GCExplainers} \label{appendix:random_gcexplainer}

As discussed in \cite{magister2021gcexplainer}, GCExplainer works by clustering a given dataset in the activation space of a given GNN, with the resulting clusters then representing the underlying concepts. In this work we use the setup from \cite{magister2021gcexplainer}, where we rely on K-means clustering as the clustering algorithm. In this case, every graph sample is assigned to one cluster in the hidden space, and is therefore part of exactly one concept (i.e., the concept representation is effectively one-hot encoded).

Consequently, our randomised versions of GCExplainer work by randomly reassigning graph samples from their original cluster to a random one, thereby representing random concept assignment.

\subsection{Concept Predictability Measurement} \label{appendix:concept_predictability}

For measuring predictability of concepts, we rely on the setup from \cite{kazhdan2020cme}. In particular, for a given concept and GNN model, we train a Logistic Regression classifier to predict the concept values from the activations of the last layer of the GNN (using other layers resulted in worse performance).

\subsection{GNN Architectures} \label{appendix:gnn_archs}

\textbf{Synthetic Baselines} For all synthetic baselines, we rely on a Graph Convolutional Neural Network (GCN) \cite{kipf2016semi}, with $3$ convolutional layers, $1$ linear layer, and global mean pooling. Further details can be found in the codebase.

\textbf{Molecular Property Prediction} For the BBBP task, we relied on a GCN with $4$ convolutional layers, $1$ linear layer, and a and global mean pooling layer.

\section{BBBP Further Results} \label{appendix:further_results}

\subsection{IA-Matrix for all samples}

\begin{figure}[h!]
    \centering
    \includegraphics[width=0.4\textwidth]{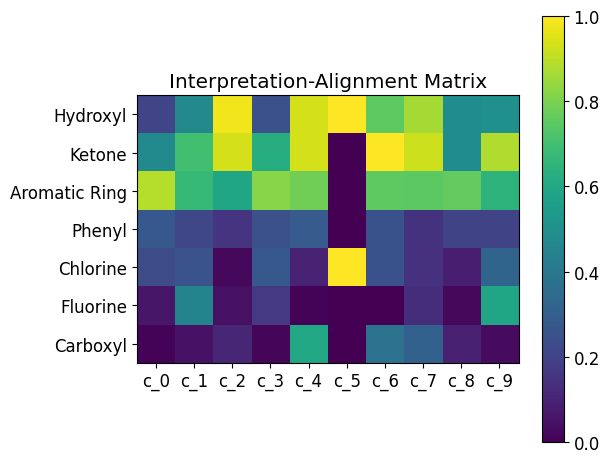}
    \caption{IA matrix computed from all samples of the BBBP dataset} 
    \label{fig:bbbp_gcexplainer_none}
\end{figure}

Figure \ref{fig:bbbp_gcexplainer_none} shows the IA matrix for both classes (i.e., for all dataset samples).

\subsection{Further Concept Samples} \label{appendix:concept_samples}

\begin{figure}[h!]
    \centering
    \includegraphics[width=0.95\textwidth]{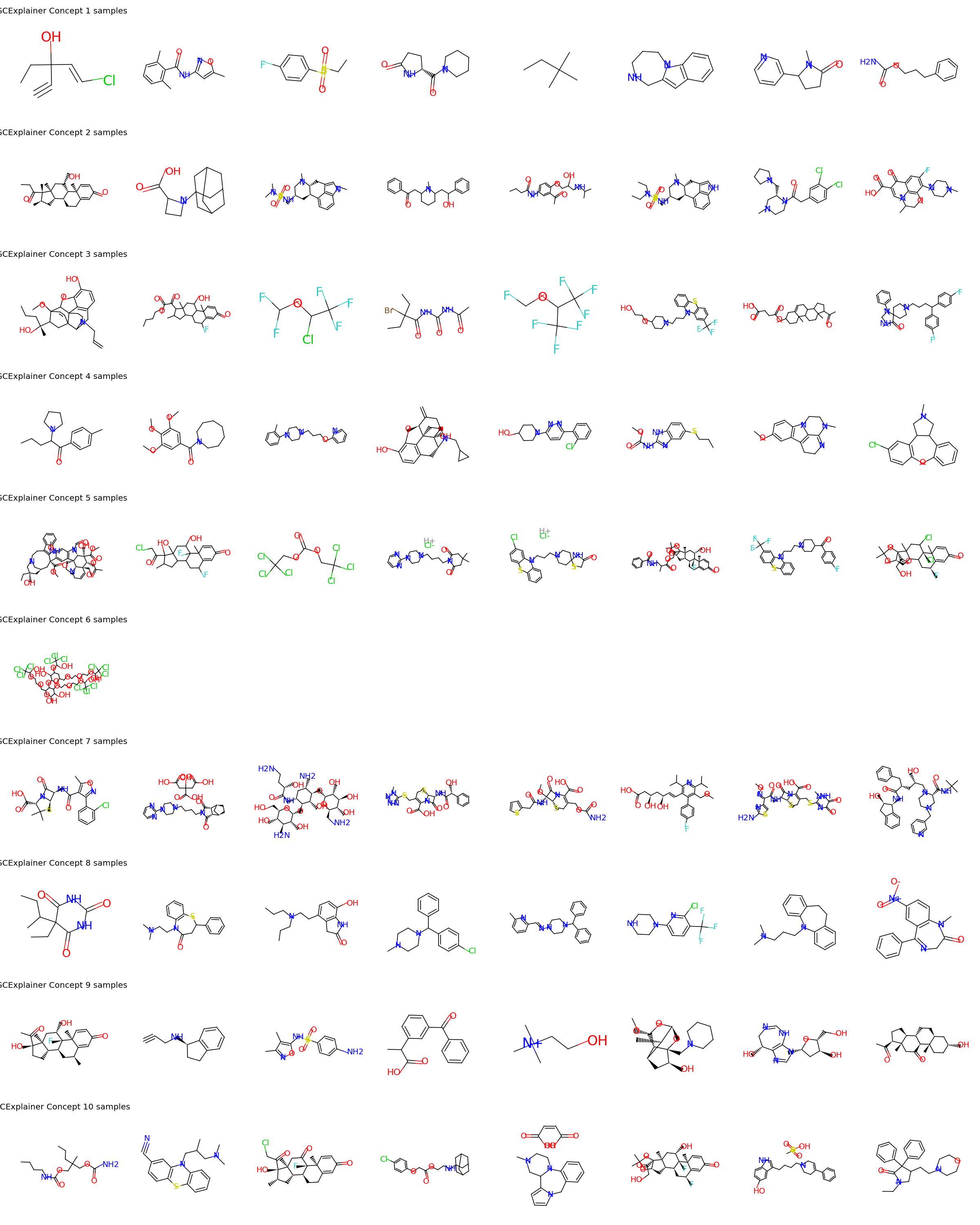}
    \caption{GCExplainer concept samples for the BBBP task, Class $1$} 
    \label{fig:bbbp_gcexplainer_samples_1}
\end{figure}

\begin{figure}[h!]
    \centering
    \includegraphics[width=0.95\textwidth]{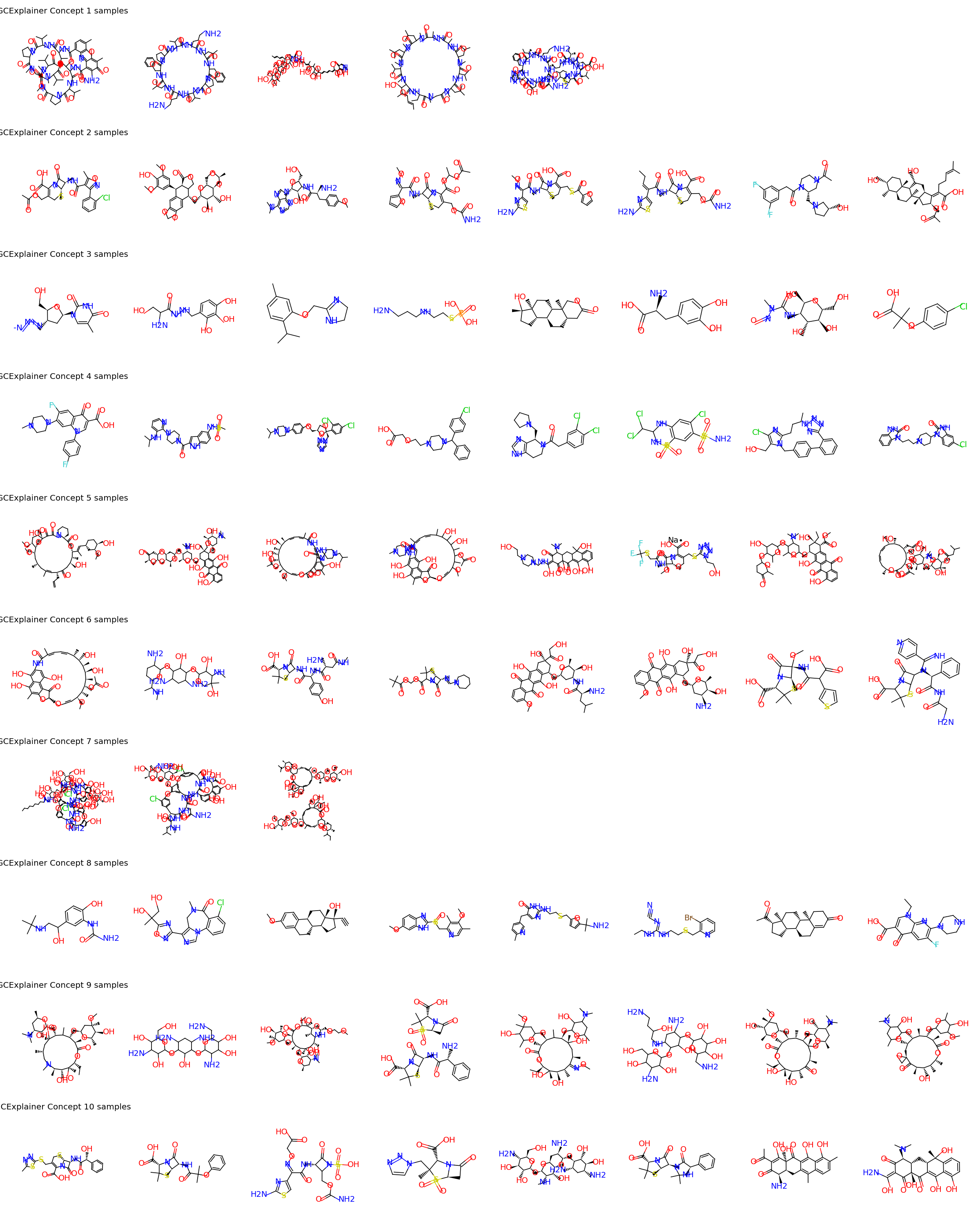}
    \caption{GCExplainer concept samples for the BBBP task, Class $0$} 
    \label{fig:bbbp_gcexplainer_samples_0}
\end{figure}

\begin{figure}[h!]
    \centering
    \includegraphics[width=0.95\textwidth]{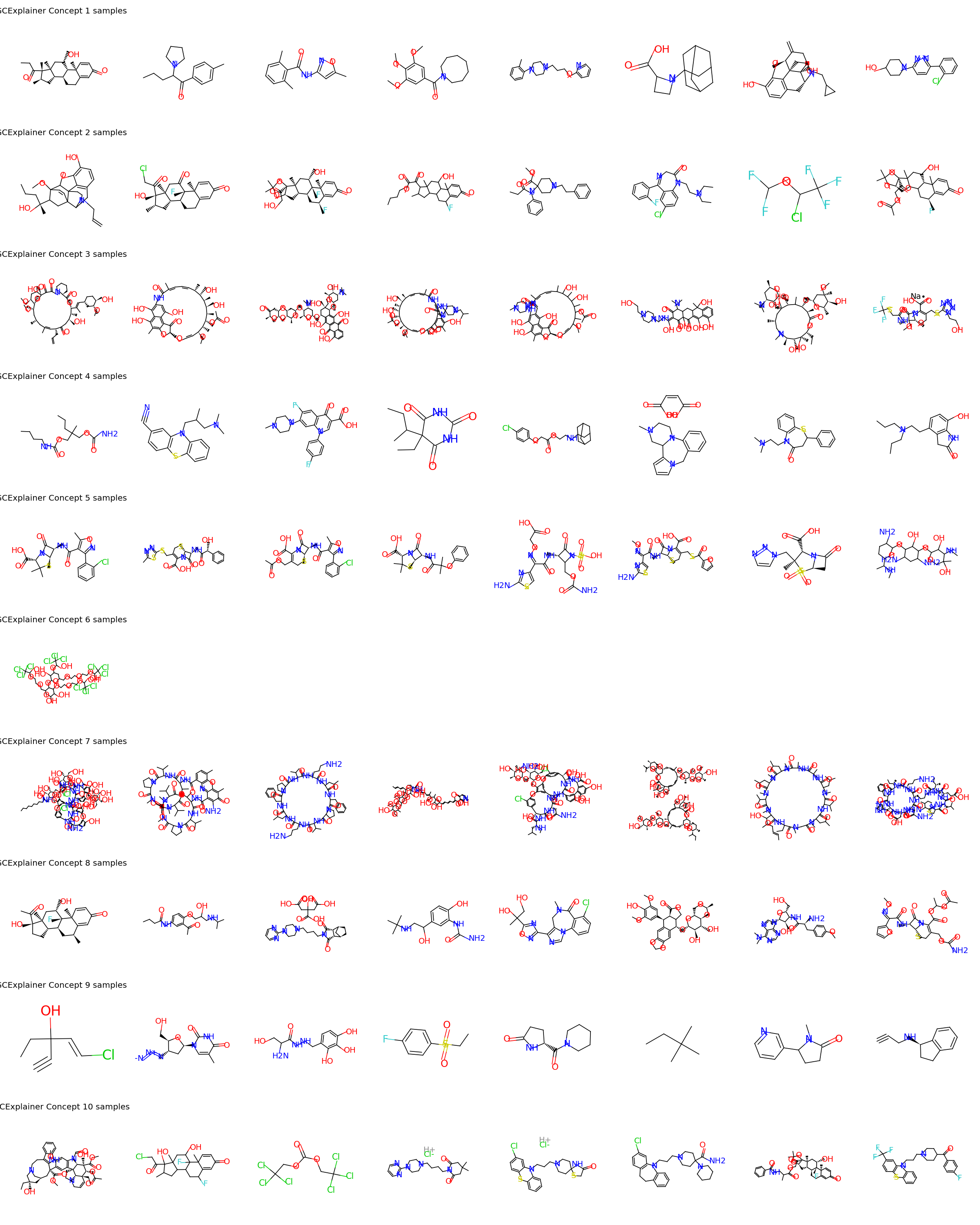}
    \caption{GCExplainer concept samples for the BBBP task, all samples} 
    \label{fig:bbbp_gcexplainer_samples_none}
\end{figure}

Figure \ref{fig:bbbp_gcexplainer_samples_1}, Figure \ref{fig:bbbp_gcexplainer_samples_0}, and Figure \ref{fig:bbbp_gcexplainer_samples_none} show the concept samples of our GCExplainer for each of the $10$ extracted concepts, for Class $1$, Class $0$, and all samples, respectively.

Overall, the IA Matrices shown in Figure \ref{fig:bbbp_matrix} serve as a compact representation of the molecules in each concept.

For instance - notice that $c_{6}$ of Class $0$ was in fact an outlier molecule extracted by GCExplainer, which indeed was predominantly consisting of \textit{Chlorine} and \textit{Hydroxyl} molecules, as demonstrated in the IA Matrix.

\newpage

\subsection{Interpretations} \label{appendix:interpretations}

\begin{figure}[h!]
    \centering
    \includegraphics[width=0.95\textwidth]{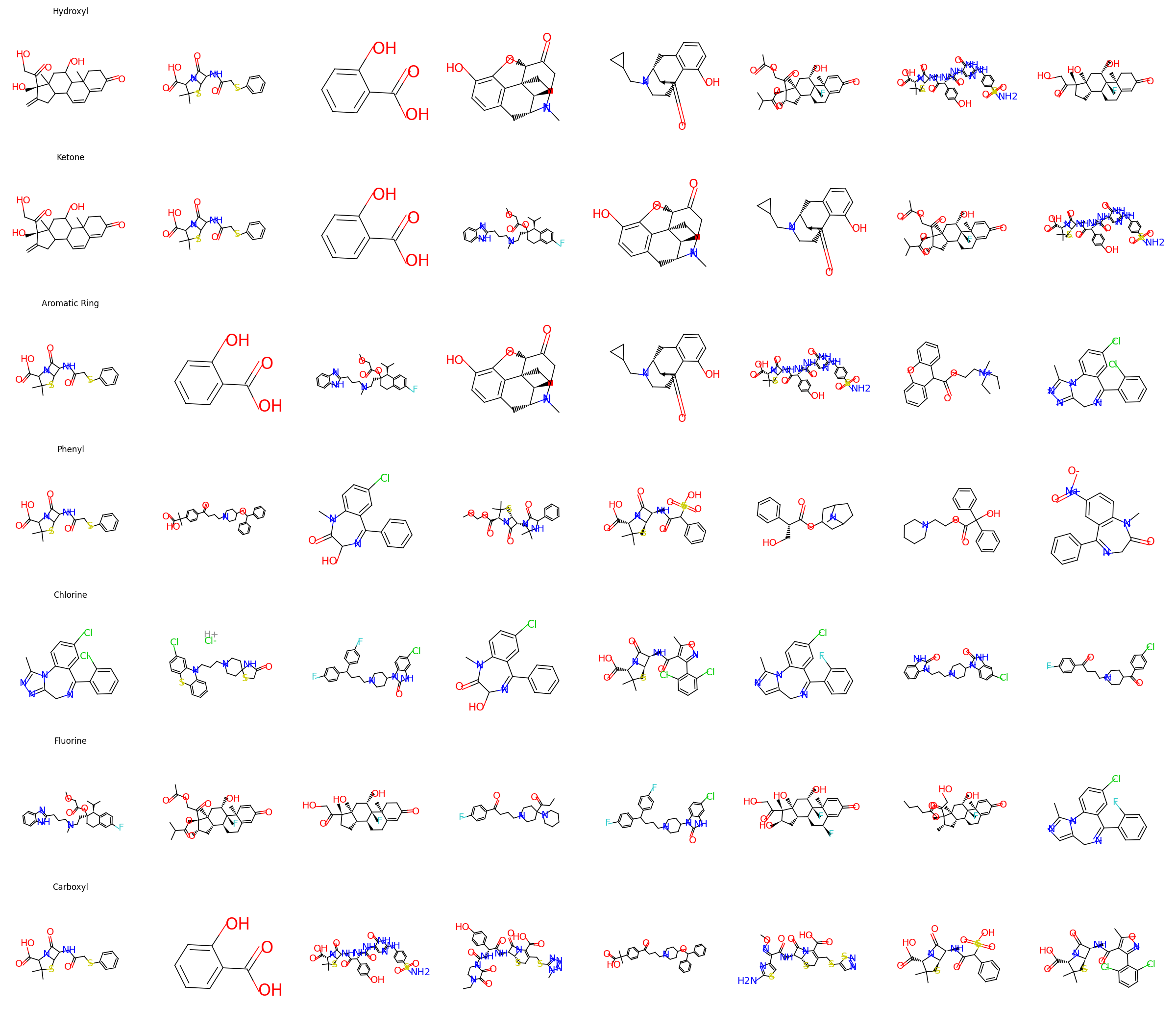}
    \caption{Graph samples for each of our interpretations used in the BBBP task} 
    \label{fig:bbbp_heuristics_samples}
\end{figure}

Figure \ref{fig:bbbp_heuristics_samples} shows a few interpretation samples for each of our $7$ heuristics.

\subsection{Varying the number of concepts} \label{appendix:bbbp_concept_var}


Below, we plot IA-Matrices for a varied number of GCExplainer concepts in the BBBP task, for Class $1.$

Importantly, GCI can be used to quickly explore and compare concept representations at varying hierarchies.

\begin{figure}[h!]
    \centering
    \includegraphics[width=0.3\textwidth]{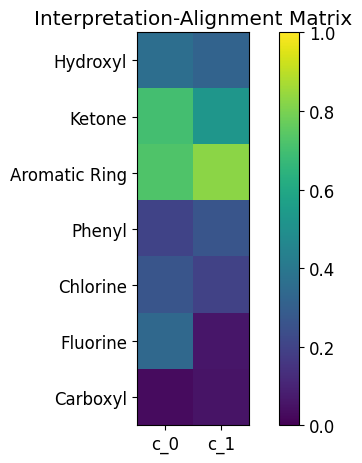}
    \caption{IA-Matrix for Class $1$, using $2$ concepts for GCExplainer}
    \label{fig:app_varying_concepts}
\end{figure}

\begin{figure}[h!]
    \centering
    \includegraphics[width=0.3\textwidth]{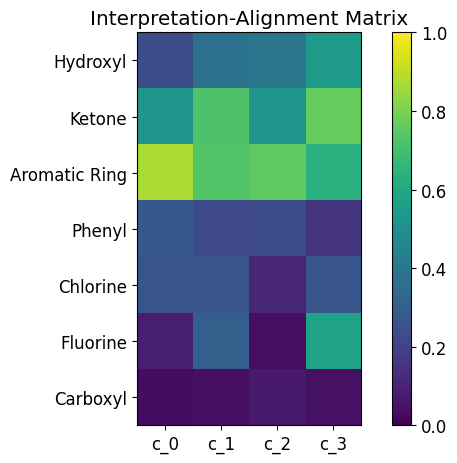}
    \caption{IA-Matrix for Class $1$, using $4$ concepts for GCExplainer}
    \label{fig:app_varying_concepts}
\end{figure}

\begin{figure}[h!]
    \centering
    \includegraphics[width=0.4\textwidth]{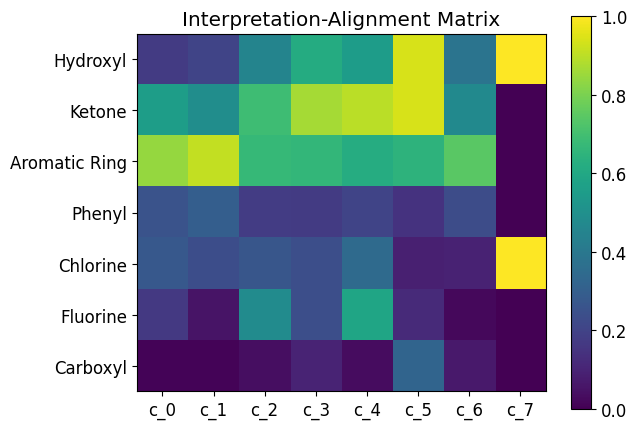}
    \caption{IA-Matrix for Class $1$, using $8$ concepts for GCExplainer}
    \label{fig:app_varying_concepts}
\end{figure}

\begin{figure}[h!]
    \centering
    \includegraphics[width=0.6\textwidth]{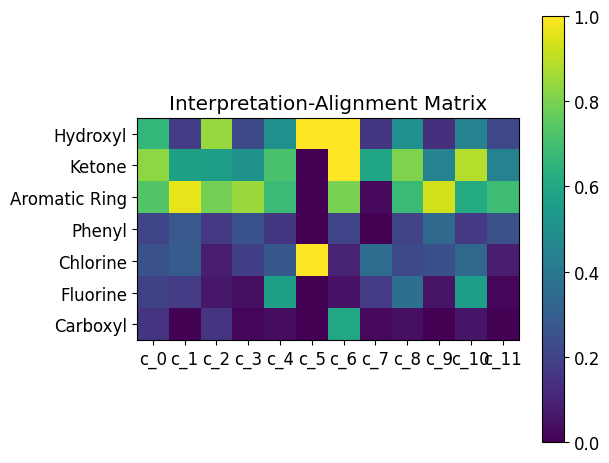}
    \caption{IA-Matrix for Class $1$, using $12$ concepts for GCExplainer}
    \label{fig:app_varying_concepts}
\end{figure}

\begin{figure}[h!]
    \centering
    \includegraphics[width=0.7\textwidth]{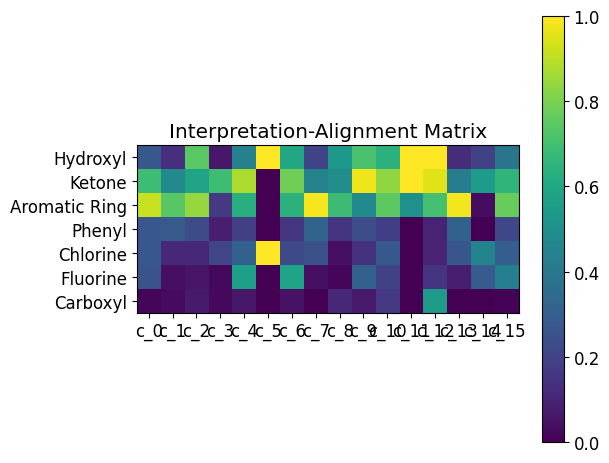}
    \caption{IA-Matrix for Class $1$, using $16$ concepts for GCExplainer}
    \label{fig:app_varying_concepts}
\end{figure}

\begin{figure}[h!]
    \centering
    \includegraphics[width=0.9\textwidth]{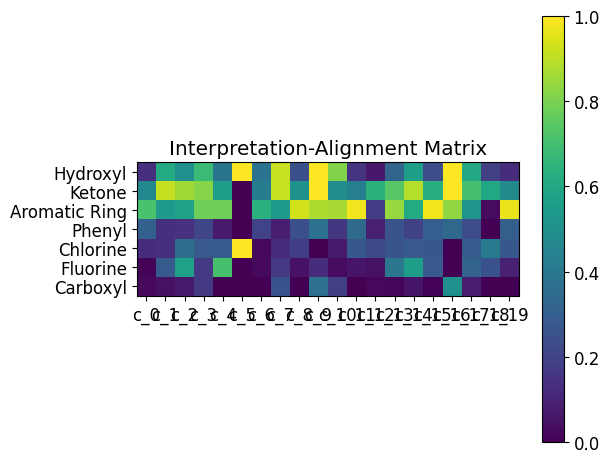}
    \caption{IA-Matrix for Class $1$, using $20$ concepts for GCExplainer}
    \label{fig:app_varying_concepts}
\end{figure}

\begin{figure}[h!]
    \centering
    \includegraphics[width=0.9\textwidth]{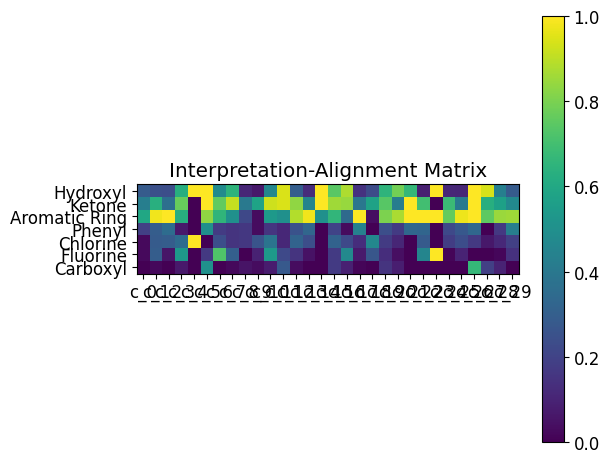}
    \caption{IA-Matrix for Class $1$, using $30$ concepts for GCExplainer}
    \label{fig:app_varying_concepts}
\end{figure}

\end{document}